\documentclass{article}

\usepackage[preprint]{corl_2026} 
\usepackage{graphicx}
\usepackage{caption}
\usepackage{xspace}
\usepackage{xcolor} 
\usepackage{caption}
\usepackage{wrapfig}
\usepackage{bbding}  
\usepackage{pifont}  
\usepackage{float}
\usepackage{amssymb}
\usepackage{xparse}
\usepackage{amsmath}
\usepackage{booktabs}
\usepackage{afterpage}

\renewcommand{\paragraph}[1]{\noindent\textbf{#1}}

\usepackage{xcolor}
\usepackage{xspace}

\definecolor{CoStreamGreen}{RGB}{0,145,95}
\DeclareRobustCommand{\ourshort}{\textcolor{CoStreamGreen}{CoStream}\xspace}

\title{CoStream: Composing Simple Behaviors for Generalizable Complex Manipulation}

\author{
  \bfseries Haonan Chen$^{*1,2}$ \quad
  Yuxiang Ma$^{*3}$ \\
  \bfseries Stephen Tian$^{2}$ \quad
  Xiaoshen Han$^{1}$ \quad
  Wenlong Huang$^{2}$ \quad
  Feiyang Wu$^{1}$ \\
  \bfseries Yunzhu Li$^{4}$ \quad
  Jiajun Wu$^{2}$ \quad
  Edward H.~Adelson$^{3}$ \quad
  Yilun Du$^{1}$ \\[4pt]
  \normalfont
  $^{1}$Harvard University \quad
  $^{2}$Stanford University \\
  $^{3}$Massachusetts Institute of Technology \quad
  $^{4}$Columbia University
}

\begin{document}
\maketitle
\begingroup
\renewcommand{\thefootnote}{*}%
\footnotetext{Equal contribution.}%
\endgroup


\begin{abstract}
Long-horizon, contact-rich complex manipulation tasks, such as seating a GPU into a PCIe slot, demand both millimeter high precision and out-of-the-box generalization to new tasks.
Existing paradigms struggle to satisfy both:
classical pipelines use brittle, task-specific interfaces to achieve high-precision control but require costly pipeline redesigns to adapt to new tasks, whereas monolithic end-to-end policies provide better generalization but lack high precision on complex, out-of-distribution tasks unless retrained with new data. Both paradigms share an implicit assumption: once a manipulation capability is acquired, it must be deployed as a rigid pipeline or monolithic whole, rather than being freely decomposed and recomposed.
In this paper, we show that complex manipulation capabilities can emerge naturally from the composition of simple, independent behaviors. Rather than deploying a monolithic policy or a rigid pipeline, we propose \ourshort, a framework orchestrating foundation models and diverse sensing modalities into multiple composable core behaviors: a semantic behavior extracting spatial constraints via foundation models; a predictive behavior forecasting trajectories by tracking keypoints in imagined videos; and a reactive behavior providing high-frequency tactile and force corrections. On a shared $SE(3)$ interface, these outputs compose by right-multiplication into a single pose command at each control step, executed by a compliant controller. 
We demonstrate \ourshort on 8 real-world tasks spanning everyday manipulation and precision assembly, with the strongest gains in contact-rich assembly and object transfer, and show robust recovery from manual perturbations during execution.
{Website:} \url{https://costream-simple.github.io}
\end{abstract}



\section{Introduction}

\afterpage{%
\begin{figure}[t]
    \centering
    \includegraphics[width=\textwidth]{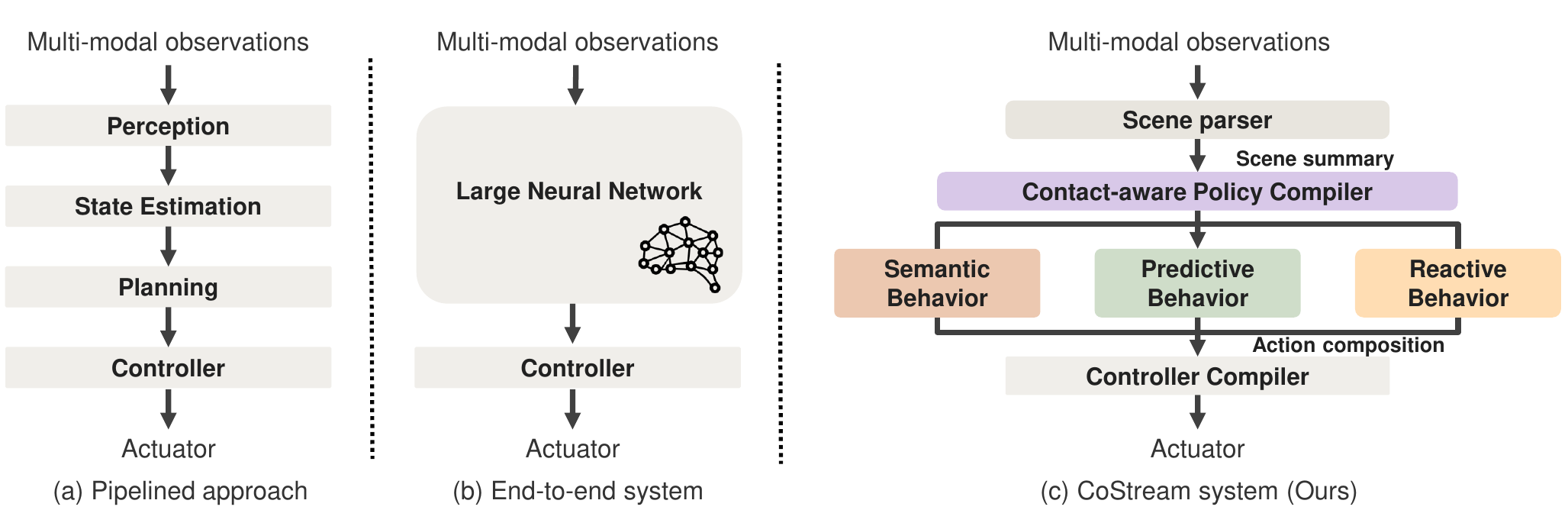}
    \caption{\small{\textbf{\ourshort: Composing Simple Behaviors for Complex Contact-Rich Manipulation.} \ourshort composes multiple sensor-grounded behaviors into a single end-effector command. A semantic behavior parses instructions with an LLM and a VLM into geometric constraints. A predictive behavior extracts a 3D reference trajectory from a video world model. A reactive behavior closes a high-rate loop from tactile and force feedback. The behaviors share an $SE(3)$ interface and compose by right-multiplication into one pose command at every control step, which a compliant controller executes while regulating contact force.}}
    \label{fig:teaser}
    \vspace{-15pt}
\end{figure}%

}

Inserting a GPU into a PCIe slot leaves less than a millimeter of clearance. A slight angular error can make the card catch on the slot rim and jam before seating. If the card slips inside the gripper mid-insertion, the entire trajectory jams. Long-horizon, contact-rich manipulation like GPU insertion requires a robot to understand the semantic geometry of a motherboard, predict a collision-free path, and physically react to sub-millimeter force deviations in real time.

To address these challenges, current research generally falls into two paradigms. The first paradigm employs classical, modular pipelines that explicitly decouple perception, state estimation, planning, and control. These pipelines are inherently rigid; adapting them to new tasks requires bespoke engineering and the manual redesign of perception and state estimation modules~\cite{shi2022robocraftlearningseesimulate, shi2023robocook, chen2023predicting}. The second relies on large monolithic policies, such as vision-language-action (VLA) models. These models attempt to absorb semantic reasoning and high-frequency contact dynamics into a single network to solve tasks like GPU insertion directly. However, these end-to-end approaches are highly data-hungry, requiring extensive new human demonstrations to generalize to every novel task~\cite{rt22023arxiv, open_x_embodiment_rt_x_2023, huang2023embodied, pomerleau1988alvinn, chi2023diffusionpolicy, ibc, octo_2023, BlackK-RSS-25, pmlr-v305-black25a, pmlr-v270-kim25c, KimM1-RSS-25, nvidia2025gr00tn1openfoundation}. Both paradigms share the assumption that complex tasks require a complex and indivisible control system.


Behavior-based robotics, a pioneering framework from the 1980s, argues that \textit{complex behavior need not arise from a single complex controller; it can instead emerge from composing simple sensory behaviors through interaction with a complex environment}~\cite{Simon1996SciencesArtificial, Brooks1986robust}. 
Inspired by this idea, we show that complex manipulation can be achieved by composing multiple simple, independent action components. Much like human motor control, breaking actions down ensures that each component remains simple. Consequently, whether a behavior is powered by a foundation model, a specialized neural network, or classical engineering, it becomes significantly easier to develop and naturally accommodates diverse sensor modalities. Building on this principle, \ourshort establishes a common SE(3) task-space interface across three behaviors: a semantic behavior that produces object-centric geometric goals, a predictive behavior that generates nominal motion from imagined video, and a reactive behavior that produces high-frequency tactile and force contact corrections. Unlike a conventional modular pipeline, \ourshort does not pass a completed plan through a sequence of fixed interfaces. Instead, these behaviors remain independent, operate at their natural rates, and compose by right-multiplication into a single pose command at every control step, which a compliant controller executes. \looseness=-1



Our contributions are threefold. First, we introduce a behavior composition interface for contact-rich manipulation that separates semantic grounding, predictive motion generation, and tactile and force reaction into independent behaviors. Second, we propose a shared SE(3) task space representation that allows each behavior to run at its natural rate and contribute only the component it is reliable for: a task frame, a nominal motion prior, or a contact correction. Third, we validate \ourshort on eight real world tasks, showing high precision assembly, manual perturbation recovery, and transfer to everyday manipulation without retraining a policy for each task.

\vspace{-7pt}
\section{Related Works}
\vspace{-3pt}

\paragraph{Foundation Models for Robotics.}
Large Language Models (LLMs) and Vision-Language Models (VLMs) have enabled robots to interpret language, reason over long-horizon tasks, and generate executable plans~\cite{saycan, rt22023arxiv, hu2023look, hong20233d, duan2024manipulate, code_as_policies}. Recent methods improve spatial precision by grounding robot actions in visual observations, using value maps~\cite{voxposer}, keypoint or affordance constraints~\cite{huang2024rekep, fangandliu2024moka, yuan2024robopoint}, iterative visual prompting~\cite{nasiriany2024pivot}, or test-time simulation~\cite{liu2025simpactsimulationenabledactionplanning}. However, these approaches still rely primarily on vision or predictive simulation, and therefore lack the real-time tactile and force feedback needed for contact-rich manipulation.

\paragraph{Tactile and Force Sensing in Robotics.}
Touch provides physical interaction cues that vision cannot capture, especially in contact-rich tasks such as cable routing, packing, and needle threading~\cite{cable_manip, robot_synesthesia, guzey2024see, ai2024robopack}. Existing tactile methods are either analytic, using tactile signals as constraints or triggers for controllers~\cite{oller2024tactile, ye2024morpheus}, or data-driven, learning skills through reinforcement learning~\cite{hu2023dexterousinhandmanipulationslender, yu2023precise}, imitation learning~\cite{hato, huang20243dvitac, bogert2024built, yu2024mimictouch, chen2025multimodalmanipulationmultimodalpolicy}, or tactile-informed MPC~\cite{tactile_mpc, ai2024robopack}. Analytic methods often require task-specific models, while data-driven methods are data-hungry and tend to specialize to the training task. In contrast, our approach uses tactile feedback to adapt object-centric behaviors online, enabling real-time adaptation and zero-shot generalization without task-specific training. \looseness=-1

\paragraph{Compositional Modeling for Robotics.}
Compositionality is central to generalizable robot behavior. Traditional modular approaches, including Task and Motion Planning (TAMP)~\cite{garrett2021tampsurvey} and layered control architectures~\cite{Brooks1986robust}, provide structure but require hand-specified interfaces and constraints. Monolithic end-to-end policies~\cite{brohan2022rt, rt22023arxiv, chen2025bimanual, zhen20243dvla} reduce manual engineering but learn entangled mappings that struggle with compositional generalization and require retraining. Recent work composes heterogeneous, factorized, or multimodal policies~\cite{wang2024pocopolicycompositionheterogeneous, liu2025flexiblemultitasklearningfactorized, chen2025multimodalmanipulationmultimodalpolicy}, while another line composes foundation models sequentially from text to visual plans and actions~\cite{ajay2023compositional, du2023video}. Our method instead composes semantic, visual, and reactive streams in parallel, avoiding both the manual engineering of TAMP and the retraining requirements of end-to-end policies.

\begin{figure*}[t]
    \centering
    \vspace{-10pt}
    \includegraphics[width=\textwidth]{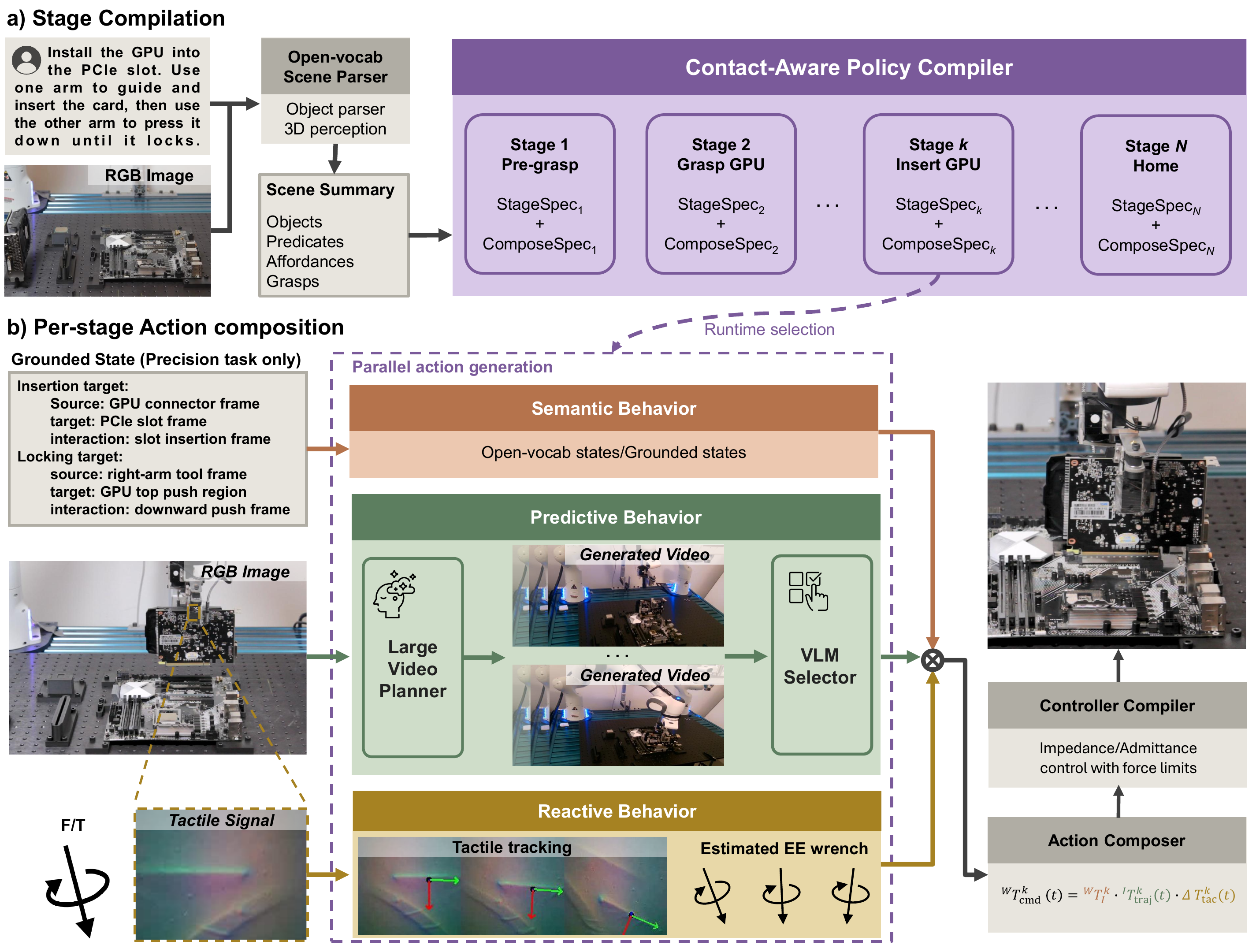}
    \vspace{-15pt}
    \caption{\small \textbf{\ourshort Architecture.} A scene parser converts language goals and observations into a Scene Summary, and a contact-aware policy compiler expands it into Stage and Composition Specifications. For each stage, the semantic behavior produces a task-frame anchor, the predictive behavior produces a nominal motion prior in that frame, and the reactive behavior produces a tactile residual and guard events. The action composer forms ${}^{W}\!T_{\mathrm{cmd}}^{k}(t)$ by right-multiplying the anchor, nominal motion, and tactile residual, while the controller compiler instantiates impedance/admittance parameters, force limits, guards, and recovery behavior for compliant execution.}
    \label{fig:methodology}
    \vspace{-20pt}
\end{figure*}

\vspace{-10pt}
\section{Methodology}
\vspace{-5pt}

\ourshort separates long-horizon contact-rich manipulation into two levels: \emph{stage compilation} and \emph{per-stage action composition} (Fig.~\ref{fig:methodology}). An open-vocabulary scene parser converts a language instruction and RGB-D observations into a \emph{Scene Summary}, which a contact-aware policy compiler expands into a sequence of typed stages. Each stage specifies which behaviors to run and how their outputs should be combined.  At runtime, the active stage runs three behaviors, semantic, predictive, and reactive, at different rates over a shared $SE(3)$ interface, which an action composer fuses into one command for compliant execution. The remainder of this section formalizes the setting and then details each level.

\vspace{-5pt}
\subsection{Problem Formulation}
\vspace{-3pt}

We consider long-horizon, contact-rich manipulation under partial observability. 
The robot is given a free-form language instruction $\mathcal{L}$ and must complete the described multi-stage task through closed-loop interaction. 
At each step $t$, the robot receives a multimodal observation $o_t = (I_t, \hat{\mathcal{X}}_t, \Psi_t)$, where $I_t$ is RGB-D, $\hat{\mathcal{X}}_t$ is an object-centric \emph{Scene Summary}, and $\Psi_t$ contains tactile and force feedback. 
The Scene Summary includes object poses, grasp frames, calibrated task frames, point clouds, affordances, symbolic predicates, and robot state. 
The policy produces actions $a_t = ({}^{W}\!T_{\mathrm{cmd}}(t), g_t)$, with target end-effector pose ${}^{W}\!T_{\mathrm{cmd}}(t) \in SE(3)$ and gripper command $g_t$.

\vspace{-5pt}
\subsection{Stage Compilation: Policy Compiler and Interface}
\vspace{-3pt}
\label{sec:interface}
Given the instruction $\mathcal{L}$ and the initial Scene Summary $\hat{\mathcal{X}}_0$, a structured policy compiler converts the task into a sequence of stages. For example, the GPU-insertion task in Fig.~\ref{fig:methodology} illustrates a multi-stage compilation, with four representative stages highlighted: pre-grasp, grasp, insert, and home. Each stage $k$ is then specified by a \emph{Stage Specification} $\mathcal{S}^k$ and a \emph{Composition Specification} $\mathcal{C}^k$ using a fixed schema.

The \emph{Stage Specification} contains the information needed to instantiate the active behaviors: the relational objective, reference frames, execution template, guard conditions, and recovery rules. We write this as $\mathcal{S}^k = (\Phi^k, F^k, E^k, G^k, R^k)$. The \emph{Composition Specification} contains the information needed to fuse behavior outputs: the composition frame, axis ownership, residual bounds, and fallback behavior. We write this as $\mathcal{C}^k = (\mathcal{F}^k, \mathcal{O}^k, \mathcal{B}^k, \mathcal{H}^k)$. These fields parameterize the fixed behavior-composition rule defined in Sec.~\ref{sec:composition}.

The specifications configure built-in behavior templates and controller profiles rather than introducing free-form code, continuous gains, or force thresholds. During execution, $\mathcal{S}^k$ parameterizes the online behaviors that produce the task-frame anchor, nominal motion, and tactile residual, while $\mathcal{C}^k$ parameterizes how these components are framed, projected, bounded, and safely composed. The controller compiler maps the selected profiles to numerical stiffness, admittance axes, residual bounds, and force/torque limits. Sec.~\ref{sec:controller} describes this deterministic controller compilation process, and Appendix~\ref{app:profiles} lists the calibrated profiles used in our experiments.

\subsection{Per-Stage Parallel Action Generation}

The three behaviors are specified by $\mathcal{S}^k$ and fused according to $\mathcal{C}^k$. Each emits only the component appropriate to its role (a frame, a motion, or a residual), rather than a complete command.

\paragraph{Semantic behavior: task-frame anchor.}
The semantic behavior emits a single anchor ${}^{W}\!T_{\mathcal I}^{k}$ from the world frame $W$ to a stage-specific task frame $\mathcal I$, updated once per stage or on reobservation. It grounds $\Phi^k$ and $F^k$ by minimizing a weighted $SE(3)$ alignment objective (Appendix~\ref{app:semantic}) over LLM/VLM outputs, keypoints, calibrated fixtures, templates, or geometric constraints. The anchor is a task frame, not a trajectory, chosen so that motion and compliance decompose along its axes~\cite{mason1981compliance, bruyninckx1996task}: this keeps axis ownership and stiffness diagonal and the tactile residual bounded. Because the anchor absorbs the object pose, the nominal motion and residual are relative and transfer across object poses and same-class instances ($SE(3)$-equivariance)~\cite{simeonov2022ndf}.

\paragraph{Predictive behavior: nominal motion.}
\label{sec:visual_traj}
The predictive behavior emits ${}^{\mathcal I}\!T_{\mathrm{traj}}^{k}(t)$, a continuous nominal motion in the task frame $\mathcal I$. A video world model~\cite{chen2025largevideoplannerenables}, conditioned on $I_t$ and the step instruction $\mathcal{L}_{\text{step}}$, samples $K$ rollouts, and a VLM critic selects the rollout $V^*$ that best satisfies the semantic and safety constraints. A 3D keypoint tracker~\cite{xiao2025spatialtracker} then lifts $V^*$ to an object-centric reference relative to ${}^{W}\!T_{\mathcal I}^{k}$, forming ${}^{\mathcal I}\!T_{\mathrm{traj}}^{k}(t)$. This is a motion prior, not a precise plan; any source that emits a task-frame trajectory can replace it.

\paragraph{Reactive behavior: tactile residual and force loop.}
The reactive behavior runs at $25$~Hz. It emits a tactile residual for the composer and drives a force loop realized by the compliant controller. It also flags contact events (slip, excess force, or loss of progress) for the runtime supervisor. The \emph{tactile residual} rejects in-hand slip: with NormalFlow~\cite{huang2025normalflow}, a GelSight Mini tracks the object's pose relative to its nominal in-hand pose, and the residual corrects this sliding online~\cite{chen2016disturbance}, keeping the semantic and predictive targets valid relative to the object. The \emph{force loop} regulates contact from the robot's estimated end-effector wrench: the controller applies a virtual spring to maintain a target force for sustained-contact stages such as wiping, and position-based admittance to bound the contact force during insertion (Sec.~\ref{sec:controller}). Appendix~\ref{app:reactive} details the tactile estimator and the compliant-control choices.

\subsection{Multi-Rate \texorpdfstring{$SE(3)$}{SE(3)} Action Composer}
\label{sec:composition}

At each controller tick the composer aligns components produced on different clocks: the anchor ${}^{W}\!T_{\mathcal I}^{k}$ is latched for the active stage, the nominal motion ${}^{\mathcal I}\!T_{\mathrm{traj}}^{k}(t)$ is interpolated, and the tactile residual $\Delta T_{\mathrm{tac}}^{k}(t)$ uses the latest in-hand sliding estimate. It first anchors the nominal motion in the world frame,
\begin{equation}
    {}^{W}\!T_{\mathrm{nom}}^{k}(t) = {}^{W}\!T_{\mathcal I}^{k}\,{}^{\mathcal I}\!T_{\mathrm{traj}}^{k}(t),
    \label{eq:nominal_composition}
\end{equation}
and then applies the tactile residual by right-multiplication,
\begin{equation}
    {}^{W}\!T_{\mathrm{cmd}}^{k}(t) = {}^{W}\!T_{\mathrm{nom}}^{k}(t)\,\Delta T_{\mathrm{tac}}^{k}(t).
    \label{eq:residual_composition}
\end{equation}
Here $\Delta T_{\mathrm{tac}}^{k}(t)$ is the rigid-body transform that cancels the latest measured in-hand sliding (Appendix~\ref{app:reactive}), expressed in the composition frame $\mathcal{F}^k$ from $\mathcal{C}^k$. The force loop is not composed here: it is realized by the compliant controller (Sec.~\ref{sec:controller}), which regulates the end-effector wrench per stage. Because ${}^{W}\!T_{\mathrm{cmd}}^{k}$ is recomputed each tick from the latest latched, sampled, and sensed inputs rather than integrated over time, drift is limited and each behavior updates at its own rate. If a component is missing or stale, the composer applies the per-stage fallback $\mathcal{H}^k$ from $\mathcal{C}^k$ (for example, holding the last valid anchor or zeroing the residual), so a dropped stream degrades gracefully rather than stalling the command.
\subsection{Controller Compiler and Compliant Execution}

\label{sec:controller}
\begin{wrapfigure}[18]{R}{0.5\textwidth}
    \centering
    \includegraphics[width=0.48\textwidth]{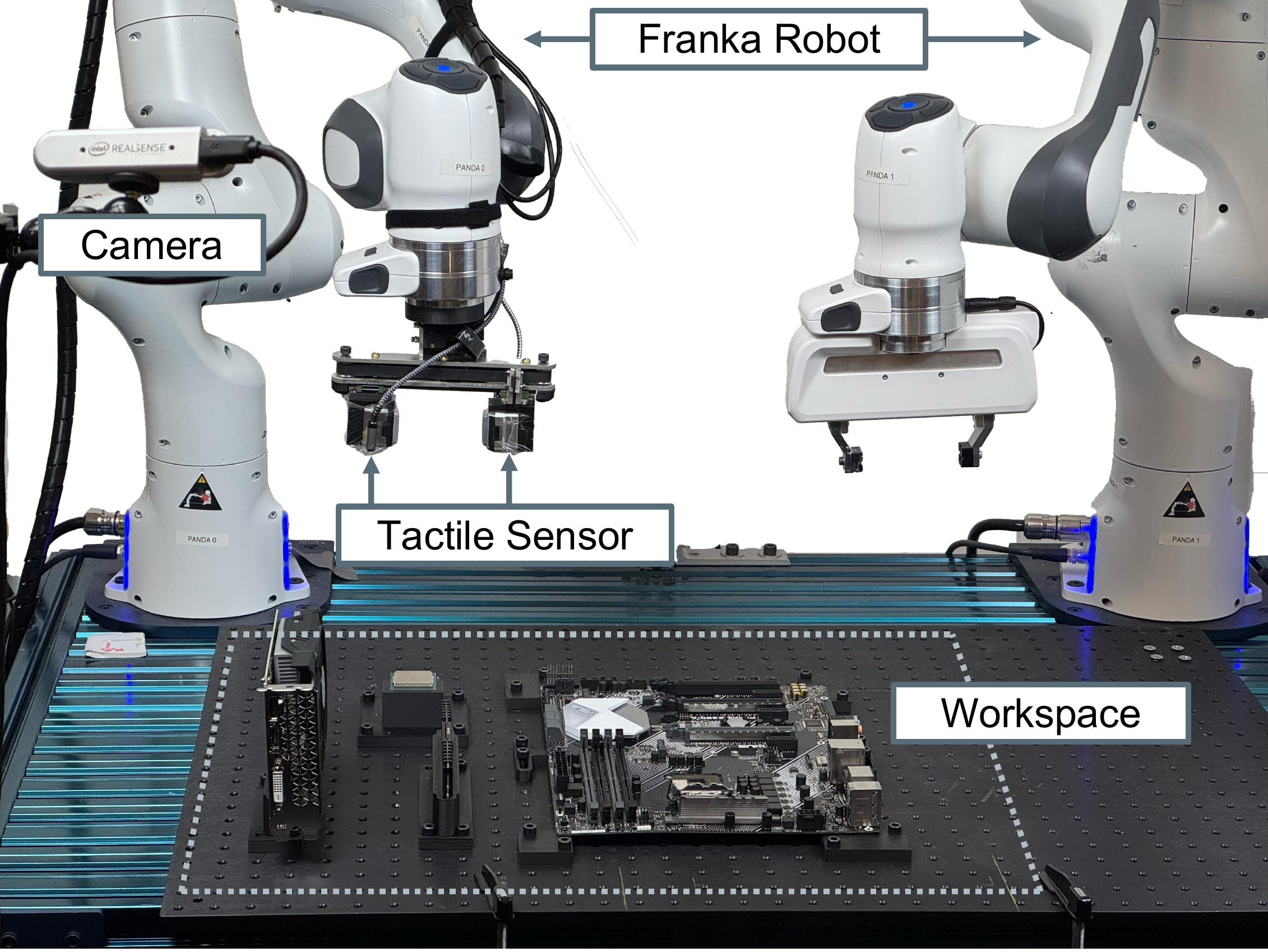}
    \caption{\small\textbf{Experimental Setup.} Two Franka Emika Panda robots; one manipulator carries a GelSight tactile sensor. The dashed outline indicates the workspace; a side-mounted camera provides visual perception.}
    \label{fig:setup}
    \vspace{-25pt}
\end{wrapfigure}

The composer outputs a task-space command; the controller compiler turns $\mathcal{S}^k$ and $\mathcal{C}^k$ into the robot-level parameters that execute it: Cartesian stiffness and damping, impedance/admittance axis selection, force/torque and residual bounds, and guard and recovery rules, instantiated from the calibrated profile library keyed by $\mathcal{S}^k$ rather than emitted by the VLM (Appendices~\ref{app:controller},~\ref{app:profiles}). In free space, the controller tracks the anchored nominal motion with high stiffness and loose contact gates. During contact, each stage applies either Cartesian impedance (a virtual spring, to maintain a target force) or position-based admittance (to bound force), with the per-axis stiffness and force regime read from the controller profile in $\mathcal{S}^k$, the owned axes $\mathcal{O}^k$ from $\mathcal{C}^k$ (along which the tactile residual is admitted), and force limits enforced from the wrench estimate. Runtime guards advance the stage on success and trigger recovery, reobservation, or replanning on excess force, no progress, slip, or pose uncertainty.


\begin{table*}[tb]
    \centering
    \caption{\textbf{Quantitative Success Rates (15 trials each).} (Left) Contact-rich assembly tasks against \textit{VoxPoser} and $\pi_{0.5}$. (Right) Everyday manipulation tasks against $\pi_{0.5}$.}
    \label{tab:results}
    \begin{minipage}{0.5\textwidth}
        \centering
        \small
        \begin{tabular}{l ccc}
            \toprule
            \textbf{Assembly Task} & \textbf{VoxPoser} & \textbf{$\pi_{0.5}$} & \textbf{Ours} \\
            \midrule
            Drill Insertion & 0/15 & 0/15 & \textbf{15/15} \\
            RAM Insertion   & 0/15 & 0/15 & \textbf{14/15} \\
            CPU Placement   & 0/15 & 0/15 & \textbf{14/15} \\
            GPU Insertion   & 0/15 & 0/15 & \textbf{15/15} \\
            \bottomrule
        \end{tabular}
    \end{minipage}%
    \begin{minipage}{0.5\textwidth}
        \centering
        \small
        \begin{tabular}{l cc}
            \toprule
            \textbf{Everyday Task} & \textbf{$\pi_{0.5}$} & \textbf{Ours} \\
            \midrule
            Turn on the Lamp           & 0/15 & \textbf{5/15} \\
            Wipe the Whiteboard        & 0/15 & \textbf{8/15} \\
            Cup $\to$ Plate            & 4/15 & \textbf{13/15} \\
            Clothes $\to$ Box          & 3/15 & \textbf{14/15} \\
            \bottomrule
        \end{tabular}
    \end{minipage}
    \vspace{-10pt}
\end{table*}

\vspace{-7pt}
\section{Experiments}
\vspace{-3pt}

To validate our framework, we conducted real-world experiments on long-horizon, high-precision robotic manipulation tasks. Our evaluation aims to address three core research questions: (\textbf{RQ1}) whether the asynchronous composition of the reactive behavior effectively rejects real-time perturbations to maintain sub-millimeter precision; (\textbf{RQ2}) whether the behavior composition architecture enables zero-shot transitions between tasks with diverse physical constraints without retraining; and (\textbf{RQ3}) whether a single set of behaviors transfers across the diverse contact regimes of CPU, RAM, GPU, and drill insertion without policy retraining or behavior-module redesign. 

\begin{figure*}[tb]
    \centering
    \includegraphics[width=\textwidth]{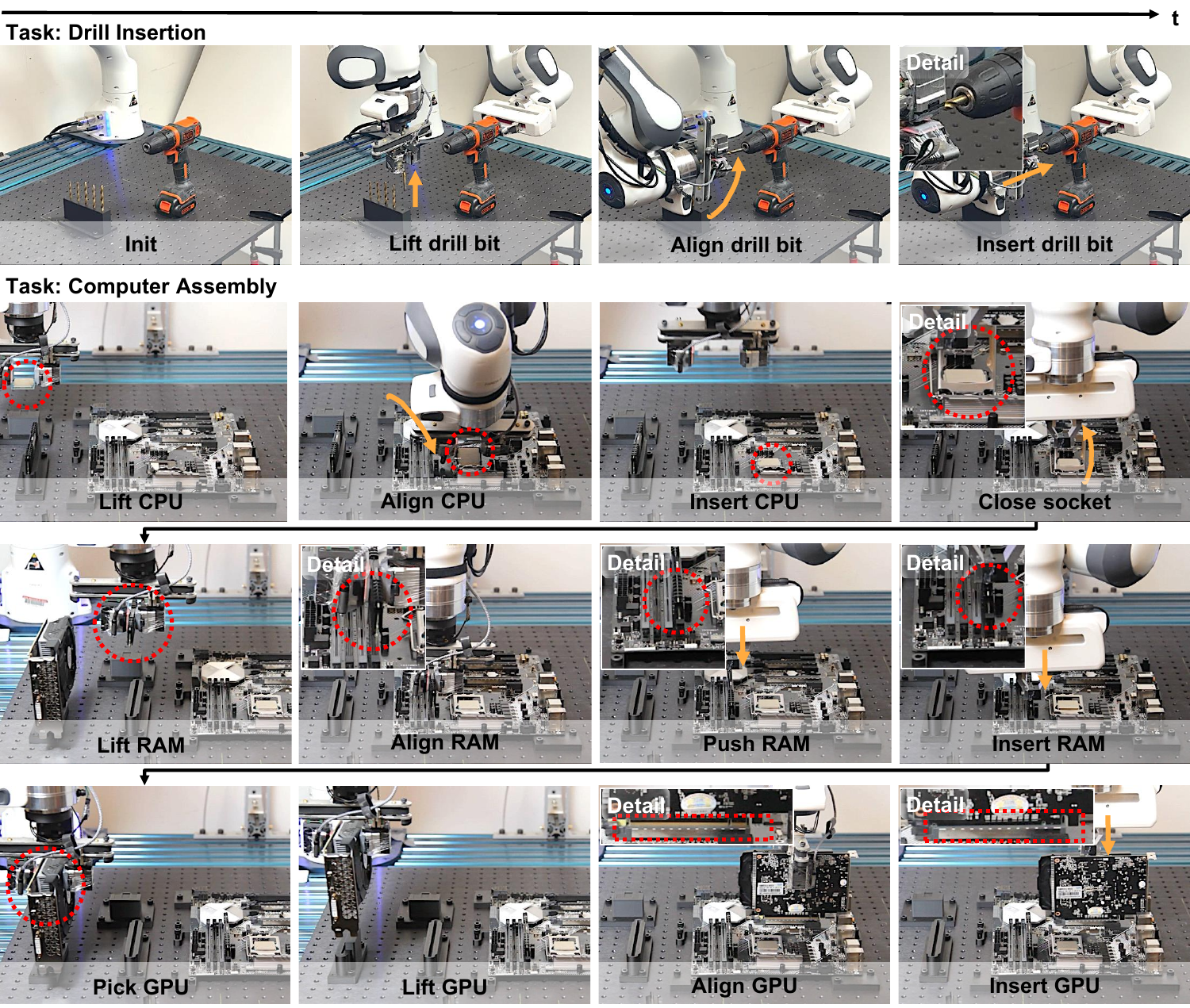}
\caption{\small{\textbf{Qualitative Rollouts of Contact-Rich Manipulation.} 
    \textbf{(Top) Drill Insertion:} The system performs high-precision insertion with a 0.5 mm clearance. The reactive behavior enables sub-millimeter compliance and corrections in real-time, maintaining contact to prevent jamming.
    \textbf{(Bottom) Computer Assembly Sequence:} The system is able to sequentially execute CPU placement, RAM insertion, and GPU installation. By updating object-centric constraints, the same architecture transitions from delicate, force-sensitive CPU handling to the high-force requirements of RAM locking without task-specific retraining.}}    \label{fig:policy_rollout}
    \vspace{-15pt}
\end{figure*}

\vspace{-15pt}
\subsection{Experimental Setup}
\vspace{-5pt}

\textbf{Hardware:} We utilize a robotic setup with two Franka Emika Panda arms. One arm is equipped with a GelSight Mini tactile sensor for precision manipulation, while the other uses a standard parallel-jaw gripper. External perception is provided by calibrated RealSense D415 cameras. Fig.~\ref{fig:setup} depicts the robots and sensor suite.

\textbf{Tasks:} We evaluate \ourshort on two task groups. The first contains four high-precision assembly tasks inspired by desktop computer assembly: drill insertion with a 0.5~mm clearance, RAM insertion requiring high-force locking, CPU placement requiring gentle flat seating, and GPU insertion requiring long-edge alignment into a PCIe slot. These tasks evaluate tight-clearance contact, robustness, and sequential assembly. The second group contains four everyday manipulation tasks: turning on a lamp, wiping a whiteboard, moving a cup to a plate, and placing clothes into a box, which evaluate broader manipulation behaviors beyond precision assembly.


\textbf{Baselines}
We compare against two representative external baselines that instantiate dominant alternatives to \ourshort: VoxPoser~\cite{voxposer}, a modular vision-language planning pipeline, and $\pi_{0.5}$~\cite{pmlr-v305-black25a}, a monolithic VLA policy used without task-specific training. These released foundation-model baselines with comparable tactile or force-feedback inputs are not available; we therefore evaluate these accessible methods under their native visual/language interfaces. These comparisons test whether current modular and monolithic foundation-model paradigms can solve the same long-horizon, contact-rich tasks under their standard assumptions. We use internal ablations to isolate the contribution of individual \ourshort behaviors and sensing streams.

\vspace{-7pt}
\subsection{Results and Analysis}

\looseness=-1
\paragraph{Quantitative success rates.}
Table~\ref{tab:results} reports real-world success across high-precision assembly and everyday manipulation tasks. In tight-clearance assembly, \ourshort succeeds consistently, while VoxPoser and $\pi_{0.5}$ fail to complete the tasks. On everyday tasks, \ourshort improves over $\pi_{0.5}$ across all settings, with larger gains on object transfer and smaller gains on lamp switching and whiteboard wiping. Since each setting uses 15 real-world trials, the numbers should be interpreted as task-level evidence rather than precise estimates of success probability. The large gap on assembly tasks nevertheless indicates a consistent qualitative difference between open-loop visual planning, monolithic VLA control, and closed-loop behavior composition.

\begin{figure*}[tb]
    \centering
    \includegraphics[width=\textwidth]{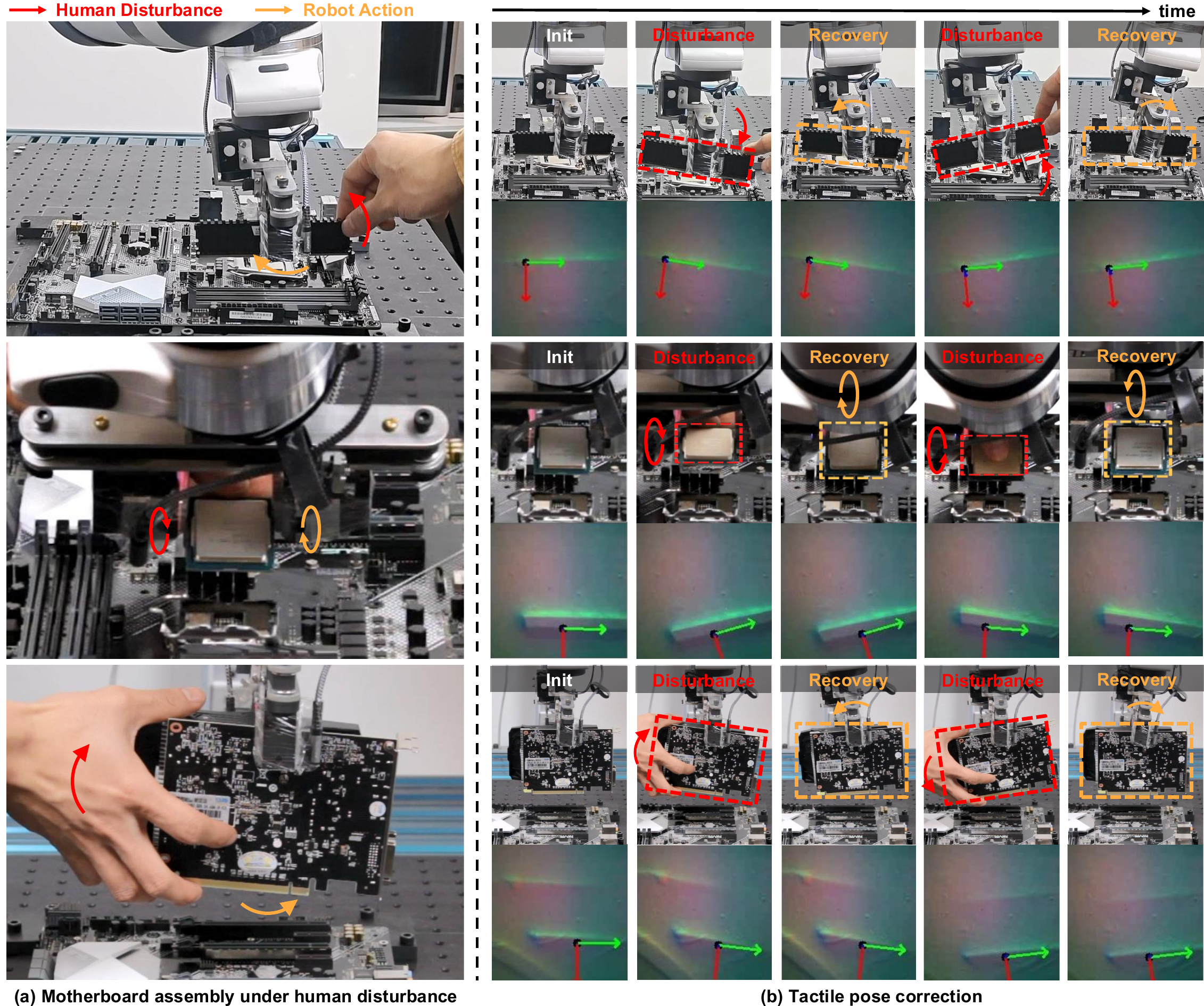}
    \caption{\small{\textbf{Robustness to Human Perturbation via Tactile Feedback.} (Left) Snapshots of the robot recovering from manual object displacement while inserting a RAM module (top), CPU (middle), and GPU (bottom). (Right) Per-component reactive alignment: each first row shows the human-induced in-gripper rotation, and each second row the raw tactile readings and derived real-time pose estimate used to realign.}}
    \label{fig:tac}
    \vspace{-15pt}
\end{figure*}

\looseness=-1
\paragraph{Failure analysis of baselines.}
VoxPoser fails before contact. Its affordance maps are constructed from a single visual snapshot and never updated; once the planner commits a trajectory, deviations from the expected geometry have no path back into the system. In our tasks the maps miss the slot orientation by several degrees, and the open-loop trajectory drives the grasped part into the rim of the hole.
$\pi_{0.5}$ fails during contact: mapping observations directly to commands, it cannot separate a contact event from perceptual ambiguity, and on contact responds with high-frequency motion that destabilizes alignment. This is exactly the failure mode \ourshort's reactive behavior eliminates: contact corrections live in a dedicated behavior with their own sensing, rate, and composition term.

\paragraph{Failure analysis of \ourshort.} The single CPU failure occurred at seating, when the chip rotated outside the tactile contact patch before the reactive term registered it; the single RAM failure occurred at locking, when the downward force briefly exceeded the compiled force-limit envelope. Both sit at the boundary of the reactive behavior's sensing or compliance envelope, not at the composition, though $N=15$ is small.

\begin{wraptable}{r}{0.5\textwidth}
    \centering
    \small
    \vspace{-20pt}
    \caption{\textbf{Ablation Study on Drill Insertion.} Removing the reactive behavior yields a significantly lower success rate, showing that contact feedback is essential for sub-millimeter precision.}
    \label{tab:ablation_drill}
    \begin{tabular}{lc}
        \toprule
        \textbf{Method} & \textbf{Success Rate} \\
        \midrule
        Ours (Full Framework) & \textbf{15/15} \\
        Ours w/o Reactive Behavior & 3/15 \\
        \bottomrule
    \end{tabular}
    \vspace{-10pt}
\end{wraptable}

\looseness=-1
\paragraph{Qualitative rollouts (RQ1 \& RQ2).} Fig.~\ref{fig:policy_rollout} shows two rollouts. In drill insertion, the bit stays centered in the 0.5~mm clearance by compliant adaptation while baselines jam. The second is a continuous CPU--RAM--GPU assembly without retraining, where only the object-centric constraints change between subtasks while the behaviors and control loops stay fixed from delicate CPU placement to high-force RAM locking.


\paragraph{Robustness to human perturbation (RQ1).} We rotate the held object inside the gripper during insertion to simulate slip. The reactive behavior detects the deviation from the reference normal map and reorients the end-effector to realign (Fig.~\ref{fig:tac}, left, across RAM, CPU, and GPU). The right panel breaks down recovery across the three components (thin RAM surfaces, a small CPU footprint, the GPU's mass).
\paragraph{Mechanism ablation (RQ1).} Without the reactive behavior on \textit{Drill Insertion} (no tactile or force feedback), success collapses (Table~\ref{tab:ablation_drill}): the controller cannot reject the angular misalignments inside the 0.5~mm clearance, succeeding only under near-perfect initial alignment.

\section{Limitations and Conclusion}
\label{sec:limitations}
\textbf{Limitations.} \ourshort inherits its perception stack: pose estimates depend on upstream modules, and visual prediction degrades when imagined rollouts drift from the scene. Our experiments target structured rigid-object manipulation with calibrated perception, reusable templates, and explicit stage specifications; deformable and articulated objects, dexterous in-hand manipulation, and ambiguous goals remain out of scope. The  reactive behavior is limited by tactile coverage and end-effector wrench estimation, so severe slip, out-of-patch contact, or inaccurate force estimates can still cause failure. Extending the composition interface to less structured settings and broader object categories is important future work.

\textbf{Conclusion.} \ourshort decomposes long-horizon contact-rich manipulation into a semantic, a predictive, and a reactive behavior, composed by right-multiplication into one pose command per control step and run by a compliant controller. The same behaviors clear four assembly tasks, recover from manual perturbation, and transfer across components without retraining, a regime that monolithic policies and modular pipelines do not reach.

\clearpage
\acknowledgments{We thank Jiayuan Mao and Pengfei Ye for helpful discussions. This work has been made possible in part by a gift from the Chan Zuckerberg Initiative Foundation to establish the Kempner Institute for the Study of Natural and Artificial Intelligence at Harvard University. This work is financially supported by Toyota Research Institute and Amazon Science Hub. }



\bibliography{example}  

\clearpage
\section{Semantic behavior: grounding solver}
\label{app:semantic}
The semantic behavior grounds the task-frame anchor ${}^{W}\!T_{\mathcal I}^{k}$ by minimizing a weighted sum of $M$ differentiable geometric residuals over $SE(3)$,
\begin{equation*}
    {}^{W}\!T_{\mathcal I}^{k\,*} = \arg\min_{{}^{W}\!T_{\mathcal I} \in SE(3)} \sum_{j=1}^M w_j\, \mathcal{J}_j({}^{W}\!T_{\mathcal I}; \hat{\mathcal{X}}_t).
\end{equation*}
Here $\hat{\mathcal{X}}_t$ is the Scene Summary, which supplies the object and fixture poses and detected keypoints used by the residuals. The active residuals $\{\mathcal{J}_j\}$ and weights $\{w_j\}$ are instantiated by the compiler from the stage objective $\Phi^k$ and grounding sources $F^k$; each $\mathcal{J}_j$ encodes one geometric relation the anchor must satisfy, such as point alignment, axis or orientation alignment, or a standoff offset. A typical point-alignment residual ties a keypoint expressed in the task frame to one on the target object,
\begin{equation*}
    \mathcal{J}_{\text{align}}({}^{W}\!T_{\mathcal I}) = \left\|{}^{W}\!T_{\mathcal I}\, \tilde{\mathbf{p}}_{\mathcal I} - {}^{W}\!T_{\text{obj}}\, \tilde{\mathbf{p}}_{\text{obj}}\right\|^2_2,
\end{equation*}
where ${}^{W}\!T_{\text{obj}}$ is the target object (or fixture) pose from $\hat{\mathcal{X}}_t$, and $\mathbf{p}_{\mathcal I}, \mathbf{p}_{\text{obj}} \in \mathbb{R}^3$ are corresponding keypoints expressed in the task frame $\mathcal I$ and the object frame, respectively, with $\tilde{\mathbf{p}}$ denoting homogeneous coordinates. We initialize with Basin Hopping, refine with Sequential Least Squares Programming (SLSQP), and warm-start across timesteps for temporal continuity.

\clearpage
\section{Reactive behavior: tactile residual and force loop}
\label{app:reactive}
\paragraph{Tactile residual (in-hand slip rejection).}
We directly track the object's pose relative to its nominal in-hand pose with NormalFlow~\cite{huang2025normalflow}, a per-frame inverse-compositional estimator~\cite{Baker2004}. Each frame solves for the warp $\Delta \xi$ that aligns the current tactile normal map to the nominal-pose reference,
\begin{equation*}
    \Delta \xi^* = \arg\min_{\Delta \xi} \sum_{\mathbf{u} \in \Omega} \left\| R(\Delta \xi)^{-1} I_{\text{obs}}(W(\mathbf{u}; \xi)) - I_{\text{ref}}(\mathbf{u}) \right\|^2,
\end{equation*}
where $\Omega$ is the contact region and $W$ is the warp induced by twist $\xi$. Because surface normal maps are scale-ambiguous in depth, we decouple the $z$ component and recover it by averaging the depth discrepancy over the contact patch,
\begin{equation*}
    \Delta z = \frac{1}{|\Omega|} \sum_{\mathbf{u} \in \Omega} \left[ D_{\text{obs}}(W(\mathbf{u}; \xi^*)) - \left( \mathbf{R}_{\xi^*} \cdot \mathbf{q}(\mathbf{u}) + \mathbf{t}_{\xi^*} \right)_z \right],
\end{equation*}
where $D_{\text{obs}}$ is the depth map from Poisson integration and $\mathbf{q}(\mathbf{u})$ is the unwarped 3D surface point at pixel $\mathbf{u}$. The recovered transform $(\Delta\xi^*, \Delta z)$ is the object's pose relative to its nominal in-hand pose, i.e., its real-time in-hand sliding. The tactile residual is the end-effector motion that compensates this sliding online~\cite{chen2016disturbance}; it is the rigid-body transform $\Delta T_{\mathrm{tac}}^{k}$ applied in Eq.~\eqref{eq:residual_composition}.

\paragraph{Force loop (compliant contact regulation).}
The force loop is realized by the compliant controller from the robot's estimated end-effector wrench, recovered from the Franka link-side joint-torque sensors mapped through the manipulator Jacobian ($F_{\mathrm{ext}} = (J^{\top})^{+}\tau_{\mathrm{ext}}$), with no wrist force/torque sensor. For stages that must sustain a contact force, such as wiping or pressing, a virtual-spring (Cartesian impedance) law regulates the measured wrench toward a target, giving approximate force regulation without dedicated force sensing. For insertion, where the goal is instead to bound contact force, position-based admittance maps the measured wrench to a bounded pose correction. We use position-based admittance rather than torque-level impedance because the end-effector carries additional cameras, tactile sensors, mounts, and cables: the resulting payload and friction mismatch degrade torque-level Cartesian tracking, whereas admittance preserves the robot's inner position servo while adding bounded, force-responsive motion that lowers the risk of jamming or damaging parts.

\clearpage
\section{Controller compilation}
\label{app:controller}
The controller compiler converts $\mathcal{S}^k$ and $\mathcal{C}^k$ into a control packet for the low-level compliant controller,
\begin{equation*}
    U^k(t) = \left({}^{W}\!T_{\mathrm{cmd}}^k(t),\, \xi_{\mathrm{cmd}}^k(t),\, K^k,\, D^k,\, A^k,\, \mathcal{B}^k,\, G^k,\, R^k\right),
\end{equation*}
where $\xi_{\mathrm{cmd}}^k(t)$ is the commanded task-space twist, $K^k$ and $D^k$ are Cartesian stiffness and damping, $A^k$ selects impedance/admittance behavior along controlled axes, $\mathcal{B}^k$ contains force, torque, residual-translation, residual-rotation, and timeout bounds, and $G^k, R^k$ define guard evaluation and fallback behavior. Stiffness is high in free space and lowered, or switched to admittance, on the axes the Composition Specification marks compliant.

\clearpage
\section{Controller and guard profiles}
\label{app:profiles}
The policy compiler chooses a per-stage controller profile and guard profile by identifier. The controller compiler then maps each symbolic level (\emph{low}, \emph{medium}, \emph{high}) to a robot-specific numerical value, calibrated once per platform from the robot, end-effector, and object-class limits with a safety margin. The LLM/VLM never produces these numbers directly: it picks a profile, not a gain or a threshold. This keeps the action space auditable and the contact behavior within calibrated safety limits. Tables~\ref{tab:controller_profiles} and~\ref{tab:guard_profiles} give a representative library; the assembly stages in this paper use the approach, insertion, press, and wiping profiles.

\begin{table}[h]
    \centering
    \small
    \caption{\textbf{Controller-profile library.} Each profile fixes the control mode and a symbolic stiffness/compliance pattern; the numerical stiffness, damping, and admittance gains are calibrated per platform.}
    \label{tab:controller_profiles}
    \begin{tabular}{@{}lp{0.17\linewidth}p{0.22\linewidth}p{0.30\linewidth}@{}}
        \toprule
        \textbf{Profile} & \textbf{Mode} & \textbf{Compliant axes} & \textbf{Stiffness / force regime} \\
        \midrule
        \texttt{free\_space} & Impedance & none & High translation and rotation \\
        \texttt{guarded\_approach} & Impedance & progress & Medium, gated on contact \\
        \texttt{compliant\_insertion} & Impedance / admittance & lateral, roll, pitch & Low lateral and roll/pitch, medium progress \\
        \texttt{high\_force\_press} & Admittance & normal & Low normal stiffness, force-controlled normal axis \\
        \texttt{surface\_wiping} & Admittance & surface normal & Force-controlled normal, free tangential motion \\
        \bottomrule
    \end{tabular}
\end{table}

\begin{table}[h]
    \centering
    \small
    \caption{\textbf{Guard-profile library.} Each profile fixes symbolic force/torque limits and a no-progress timeout; numerical thresholds are calibrated with a safety margin.}
    \label{tab:guard_profiles}
    \begin{tabular}{@{}lp{0.22\linewidth}p{0.16\linewidth}p{0.24\linewidth}@{}}
        \toprule
        \textbf{Profile} & \textbf{Normal-force limit} & \textbf{Torque limit} & \textbf{No-progress timeout} \\
        \midrule
        \texttt{delicate\_placement} & Low & Low & Short \\
        \texttt{rigid\_insertion} & Medium & Medium-low & Medium \\
        \texttt{high\_force\_locking} & High (bounded) & Medium & Short \\
        \texttt{surface\_wiping} & Medium band & Medium & Disabled (sustained contact) \\
        \bottomrule
    \end{tabular}
\end{table}

\clearpage
\section{Perception module}
\label{app:perception_software}
The Scene Summary $\hat{\mathcal{X}}$ consumed by the policy compiler (Sec.~\ref{sec:interface}) is produced by an open-vocabulary perception stack. We adopt the two-branch design of TiPToP~\cite{shen2026tiptop}: a 3D vision branch recovers metric geometry, and a semantic branch grounds the instruction to objects. We then upgrade both branches and extend their fused output, as detailed below.

\paragraph{3D vision branch.}
From the calibrated RGB-D streams, FoundationStereo~\cite{wen2025foundationstereo} estimates a dense scene point cloud, and M2T2~\cite{yuan2023m2t2} predicts candidate $6$-DoF grasps from it. Masking the dense cloud with the semantic branch below yields per-object point clouds, and convex-hull reconstruction completes object geometry, following TiPToP~\cite{shen2026tiptop}. The resulting meshes support collision checking and grasp-frame estimation.

\paragraph{Semantic branch.}
Gemini Robotics-ER 1.6~\cite{team2025gemini, deepmind2026geminiER} detects task-relevant objects and grounds the instruction $\mathcal{L}$ to symbolic goal propositions. SAM 3~\cite{carion2025sam} then lifts these detections to per-object masks through concept prompts that index into the point cloud. This branch supplies the object identities, language-grounded targets, and symbolic predicates of $\hat{\mathcal{X}}$.

\paragraph{Scene Summary construction.}
Fusing the two branches yields an object-centric record per task-relevant object: pose, completed mesh, and candidate grasps. We annotate each record with three further fields. \emph{Affordances}, such as an insertable bore, a slot, or a graspable edge, are attached to specific object parts. \emph{Interaction and task frames} are attached to those affordances, and seed the semantic behavior's anchor ${}^{W}\!T_{\mathcal I}^{k}$. \emph{Available geometry sources}, such as calibrated fixtures, object-centric templates, or estimated poses, are tagged per object. These tags let the compiler judge which frames are reliable for a high-precision stage. Together, the three fields let $\hat{\mathcal{X}}$ compile into Stage and Composition Specifications without a separate grounding pass.

\paragraph{Relation to TiPToP.}
Our stack differs from TiPToP~\cite{shen2026tiptop} in both components and purpose. TiPToP pairs Gemini Robotics-ER 1.5 with SAM-2, and feeds the resulting meshes and symbolic goals to a TAMP planner. We instead use Gemini Robotics-ER 1.6, whose spatial and physical reasoning improves over 1.5~\cite{team2025gemini, deepmind2026geminiER}. We also replace SAM-2 with SAM 3, whose concept-promptable segmentation grounds open-vocabulary objects more reliably~\cite{carion2025sam}. More importantly, our added annotations let $\hat{\mathcal{X}}$ compile directly into Stage and Composition Specifications for contact-rich behavior composition. The same representation in TiPToP instead feeds a motion planner.

\paragraph{Limitations.}
The stack builds the Scene Summary once from the initial views, so errors in the estimated poses or frames propagate to every stage. The reactive behavior corrects only contact-observable deviations on its owned axes, not global perception errors. High-precision frames therefore still rely on calibration or object-centric templates rather than purely open-vocabulary estimation. Convex-hull completion yields coarse geometry, sufficient for collision checking but not for fine reasoning on concave parts. Finally, the semantic branch depends on a proprietary model (Gemini Robotics-ER 1.6) served through an API, which constrains reproducibility, offline use, and latency.

\clearpage
\section{Sensing hardware and calibration}
The sensing stack consists of two calibrated RealSense D415 cameras mounted at fixed external viewpoints and a GelSight Mini tactile sensor on one of the two Franka end-effectors. Camera extrinsics are obtained by an ArUco-board hand-eye calibration repeated whenever the workspace is reconfigured. Object pose estimates that feed the semantic and predictive behaviors are produced by an upstream perception module adapted from prior work; downstream behaviors treat these poses as part of the externally provided Scene Summary $\hat{\mathcal{X}}_t$, and the composition interface is agnostic to the specific pose source. The tactile stream is read at 25~Hz (the GelSight Mini readout rate), and the surface normal map used by the reactive behavior is computed per frame via Poisson integration on the raw gel image. The reactive behavior consumes only the tactile stream and does not depend on the external cameras during contact, which is what allows the corrective loop to continue when an object is occluded by the gripper.

\clearpage
\section{Visual Trajectory Generation}
\begin{figure*}[h]
    \centering
    \includegraphics[width=\textwidth]{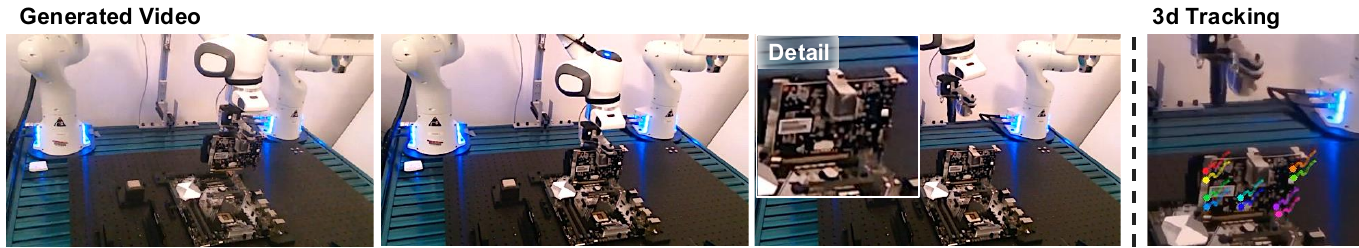}
    \caption{\small{\textbf{Visual Trajectory Generation and Keypoint Tracking.} The left part illustrates the video frames generated by the world model (e.g., GPU insertion), while the right figure depicts the extracted 3D keypoint tracks. These imagined futures provide a high-fidelity motion prior for deriving smooth, object-centric trajectories.}}
    \label{fig:video_gen}
\end{figure*}

\end{document}